\documentclass{article} % For LaTeX2e
\usepackage{iclr2025_conference,times}
\usepackage{graphicx}
\usepackage{tabularx} % For flexible table widths
\usepackage{booktabs} % For better lines in tables
\usepackage{multirow} % For multi-row cells
\usepackage{caption}   % To customize captions
\usepackage[T1]{fontenc}

\usepackage{hyperref}
\usepackage{url}

\iclrfinalcopy 
\title{RisingBALLER: A Player is a Token, a Match is a Sentence–A Path Towards a Foundational Model for Football Players Data Analytics}

\author{Akedjou Achraff ADJILEYE \\
Paris, FRANCE \\
\texttt{adjileyeb@yahoo.com} \\
}

\begin{document}

\maketitle

\begin{abstract}
In this paper, I introduce RisingBALLER, the first publicly available approach that leverages a transformer model trained on football match data to learn match-specific player representations. Drawing inspiration from advances in language modeling, RisingBALLER treats each football match as a unique sequence in which players serve as tokens, with their embeddings shaped by the specific context of the match. Through the use of masked player prediction (MPP) as a pre-training task, RisingBALLER learns foundational features for football player representations, similar to how language models learn semantic features for text representations. As a downstream task, I introduce next match statistics prediction (NMSP) to showcase the effectiveness of the learned player embeddings. The NMSP model surpasses a strong baseline commonly used for performance forecasting within the community. Furthermore, I conduct an in-depth analysis to demonstrate how RisingBALLER’s learned embeddings can be used in various football analytics tasks, such as producing meaningful positional features that capture the essence and variety of player roles beyond rigid x,y coordinates, team cohesion estimation, and similar player retrieval for more effective data-driven scouting. More than a simple machine learning model, RisingBALLER is a comprehensive framework designed to transform football data analytics by learning high-level foundational features for players, taking into account the context of each match. It offers a deeper understanding of football players beyond individual statistics.
\end{abstract}

\section{Introduction}\label{sec: Introduction}

In recent years, the field of machine learning has been revolutionized by the introduction of the transformer architecture [1], which initially gained prominence in natural language processing (NLP) with models like BERT [2], RoBERTa [3], and more recently, the widespread use of large language models (LLMs). These models, often trained on seemingly simple tasks such as next token prediction or masked token prediction, have demonstrated remarkable performance in learning high-level features that effectively represent each word and model language intricately. They are capable of learning nuanced representations of the multiple meanings a word can have depending on its context. This led to the era of language foundation models—models trained on large text corpora that provide foundational representations of words, which are then used for various general and domain-specific tasks such as text classification, document retrieval, hate speech filtering or zero-shot question answering through conversational agents. The concept of foundation models has also emerged in computer vision, with models like CLIP [4] and DiNO [5] trained on large datasets of images and text to learn high-level image representations, central for multiple vision tasks such as object detection and tracking, image captioning and video understanding.

Meanwhile, the football analytics community still heavily relies on handcrafted representations for players, which require extensive feature engineering which is often time-consuming, difficult to reproduce, and challenging to scale, making it impractical for long-term and large-scale usage. For instance, [7] constructed a distance metric using a dataset of 3,003 players from the 2014-2015 season across the top 8 European leagues by manually crafting 109 performance variables. This distance metric was then used to identify similar players and cluster players with comparable profiles. Although this approach yielded good results, the learned features are specific to that particular season and cannot be applied to data spanning multiple seasons. Similarly, [8] proposes Transfer Portal, a framework for predicting a player's performance after a potential transfer from one club to another. Although the primary goal is not to retrieve similar players, the method involves representing players by aggregating their per 90-minute stats over a fixed number of matches per season, making them impossible to use for general player-level tasks. Other works [9, 10] in the literature also rely on handcrafted features based on player statistics over a certain number of matches for various football data analytics tasks.

The success of the foundation models in NLP and computer vision raises intriguing questions: \textbf{Can the concept be applied to other fundamental domains, such as football analytics, to learn high-level foundational representations for football players? and teams? If so, how?}

Football, much like text or visual scenes, is a language of actions spoken by players throughout a match and expressed through performance data and metrics [6]. Each player uniquely contributes to the narrative and outcome of the match. Thus, it becomes clear that the same principles that underlie text or vision foundation models can be applied to the "language" of football to create high-level player representations that can be used later in various football analytics tasks.

This paper explores this idea by proposing a novel method that leverages the transformer architecture to model the "football language" spoken by players at each match. In this approach, each player serves as a token—the fundamental unit for training a transformer network. Their representations are randomly initialized and then dynamically adjusted by the transformer model as it is trained on match data. Specifically, by assigning unique IDs to players within a set of football matches, I construct a vocabulary of encodable tokens. These tokens are enriched with player statistics from the match to serve as temporal positional encodings (TPE), along with spatial representations of player positions and team representations. Drawing further inspiration from text language modeling, the model I present is trained on multiple football matches, with each match represented as a sequence of players. I propose masked player prediction (MPP) as an effective pre-training task for this model to learn robust player features.

The model, named \textbf{RisingBALLER}—a term used to describe a promising young player in football—aims to be a pioneering effort toward the adoption of the foundation model paradigm in football analytics research. RisingBALLER aspires to bring the same level of utility and scalability achieved by NLP and computer vision foundation models to football data analytics. The key contributions of the paper are as follows:
\begin{itemize}
    \item I introduce RisingBALLER, a transformer-based model that treats football matches as sequences of players, trained using masked player prediction (MPP) to learn effective player feature representations.
    \item I propose Next Match Statistics Prediction (NMSP) as a downstream task for RisingBALLER. The results show that the fine-tuned model outperforms a strong and common baseline in the community, which forecasts next match statistics as the average of the previous five matches' statistics.
    \item I conducted a comprehensive architecture search for MPP pre-training, considering the limited data available, and demonstrated its crucial role in achieving strong performance in NMSP. 
    \item Additionally, I performed an in-depth analysis of the player and position embeddings learned by RisingBALLER, showcasing their applications, particularly in understanding players' tactical roles, data-driven team building, and recruitment strategies.
\end{itemize}

To the best of my knowledge, this is the first public work leveraging the foundation model paradigm to learn high-level features for representing football players. The code will be released\footnote{https://github.com/akedjouadj/risingBALLER} to encourage further research and development in this field.

The remainder of this paper is structured as follows: Section \ref{sec: Methodology} introduces the dataset and model architecture; Section \ref{sec: Experiments and Results} outlines the MPP and NMSP training processes and results; Section \ref{sec: RisingBALLER applications, the power of learnable embeddings} explores the potential of the embeddings learned by the model; Section \ref{sec: Computational resources/Technical details} provides technical details on the training process and computational resources used; and Section \ref{sec: Conclusion and Future Works} concludes the paper and outlines future research directions. Additional ablation studies and results from the embeddings analysis are provided in the Appendix.

\section{Methodology}\label{sec: Methodology}
\subsection{Dataset}\label{subsec: Dataset}
I used event data provided for free by StatsBomb for all matches in the 2015-2016 season across the top 5 European leagues: the English Premier League, La Liga, Bundesliga, Serie A, and Ligue 1. For each match, there is a dataframe containing between 3,500 and 4,000 rows, each representing an action that occurred during the match from start to finish, with each action described by numerous features, including the player involved.

I convert each match event dataframe into a player statistics dataframe, where each row represents a player in the squad of the two teams that played the match. This dataframe includes identification features such as the player's team, position, league, a mask indicating whether the player actually participated in the match (either as part of the starting 11 or as a substitute), and various player statistics from the match, including passing stats, offensive stats, defensive stats, and goalkeeper stats. If a player did not participate in the match, all their statistics are set to 0.

I collected 39 statistics from the StatsBomb data to represent each player. These include:
\begin{itemize}
    \item 10 statistics related to passing: pass total, pass cross, pass cut back, pass shot assist, pass goal assist, pass no touch, pass interception, pass incomplete, pass offside, pass through ball.
    \item 9 statistics related to shots: shot total, shot statsbomb xg, shot corner, shot free kick, shot open play, shot penalty, shot saved, shot off target, shot blocked, shot goal.
    \item 3 statistics related to interceptions: interception total, interception won, interception lost. 
    \item 3 statistics related to dribbles: dribble total, dribble complete, dribble incomplete.
    \item 5 statistics related to fouls: foul won total, foul won penalty, foul committed total, foul committed penalty, foul committed yellow card, foul committed red card.
    \item 3 statistics related to goalkeeping: goalkeeper goal conceded, goalkeeper save, goalkeeper shot faced. 
    \item additionally: block total, clearance total, ball recovery total and counterpress total.
\end{itemize}

For the next match statistics prediction task, matches are sorted by their kickoff date. Each of these statistics is then aggregated into sums, means, and standard deviations to model the quantity of matches played, the mean, and the variance of performance—both from the beginning of the season up to the match being predicted and over the last 5 matches prior. This process generates 234 statistical variables to model each player for each match in the dataset.

In total, I generated data for 1,792 matches, representing 2,600 unique players (the player vocabulary) from 98 unique teams. For the masked player prediction (MPP) task, I added an ID to represent the masking token and an ID for the padding token, allowing all inputs to have the same dimensions (a fixed number of players per match) for batch computation. To compute team-level stats for the next match statistics prediction (NMSP), I summed all the statistics for all the players in the team’s squad.

\subsection{The Model Architecture}\label{subsec: The Model Architecture}
\begin{figure}[h]
\begin{center}
\includegraphics[width=\textwidth]{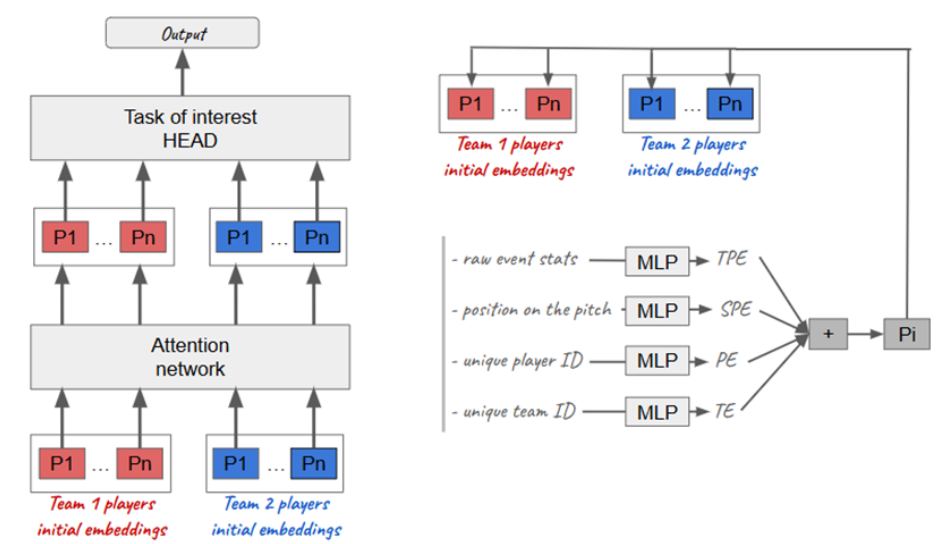}
\end{center}
\caption{Architecture of RisingBALLER—each player in the match dataset is treated as a token, with a unique ID embedded into a D-dimensional feature vector (PE). This ID is combined with additional embeddings representing the player's spatial position on the field (SPE) and team affiliation (TE). The player's match event data is also projected into the same D-dimensional space, serving as temporal positional embeddings (TPE). These four vectors are then combined to initialize the player representation before being fed into the attention network.}
\label{fig: 1}
\end{figure}

RisingBALLER is a transformer-based model specifically designed to represent football players using match event data. The core idea behind the architecture is to treat each player participating in a match as a token, leveraging the contextual nature of football matches by encoding match-related features into dense representations that can be utilized for various downstream tasks.
The input to the RisingBALLER model consists of sequences corresponding to the players from both teams participating in a match, as illustrated in Figure \ref{fig: 1}. For each player, the following components are included:
\begin{itemize}
    \item \textbf{Unique Player ID Embedding (PE):} Each player is assigned a unique ID, which is embedded into a D-dimensional space using a Multi-Layer Perceptron (MLP) on a one-hot encoding vector representing the player's ID within the player vocabulary. This embedding serves as the primary representation of the player within the model.
    \item \textbf{Spatial Positional Embedding (SPE):} The player's position on the field is critical for understanding their role and contribution to the game. The player’s position is encoded into a D-dimensional vector, also using an MLP on a one-hot encoding vector of the position type. This encoding captures the spatial context of the player within the match.
    \item \textbf{Temporal Positional Encoding (TPE):} The match event statistics of each player, which vary from match to match, are embedded into the same D-dimensional space. This encoding provides a temporal representation of the player's form and performance during the match. The raw event data is projected into the embedding space through an MLP.
    \item \textbf{Team Affiliation Embedding (TE):} To distinguish between players from different teams, a team ID embedding is introduced. Like the player and position IDs, each team ID is embedded into a D-dimensional space using an MLP. This embedding provides context regarding the player’s team affiliation.
\end{itemize}

These four components—Player ID (PE), Spatial Position (SPE), Team Affiliation (TE), and Temporal Positional Encoding (TPE)—are combined through element-wise addition to produce the initial player representation in a unified D-dimensional space. The initialized player representations tensor for both teams $X_{init}$, of shape $[N_{players}, D]$, is then fed into a transformer-based attention network. The transformer architecture employed follows the standard structure introduced in [1]. The output tensor $X_{out}$ with the same shape as $X_{init}$ contains match-contextualized representations for the players and is used for the downstream task of interest.

For the Masked Player Prediction (MPP) task, $X_{out}$ is projected through an MLP into a V-dimensional space, and a softmax activation function is applied to compute the probability of the masked player being one of the players in the vocabulary.

For the Next Match Statistics Prediction (NMSP) task, all player representations are flattened, and the resulting one-dimensional tensor of shape $[N_{players}*D]$ is projected through an MLP into a $2*N_{stats}$ dimensional space to predict the $N$ statistics of interest for each of the two teams.

\section{Experiments and Results}\label{sec: Experiments and Results}
\subsection{Pre-Training with Masked Players Prediction (MPP)}\label{subsec: Pre-Training with Masked Players Prediction (MPP)}
Transformer models learn robust player representations through pretext tasks, which are crucial for achieving strong performance on downstream tasks. Inspired by the masked token prediction used in language modeling, I initially trained RisingBALLER using Masked Players Prediction (MPP). The goal of MPP is to enable the model to learn some prior relationships between players. The MPP task is straightforward: for each match, 25\% of the input players are randomly masked, and the model is trained to predict these masked players.

I trained two models with one transformer layer each, using embedding dimensions of D=64 and D=128 respectively, across the 1,792 processed matches in the dataset. To enhance the model's generalization capabilities given the relatively small dataset size, I augmented the data by creating 10 different MPP inputs per match, each with different players masked. This resulted in approximately 18,000 input samples. The maximum sequence length per match was set to 80, and padding tokens were added if needed. I used 80\% of the input samples for training and 20\% for validation. The total number of tokens used for training was approximately 1.14 million, with a vocabulary size of 2,602, including the 2,600 unique players, the masking token, and the padding token.

The two models were trained with a batch size of 256, a learning rate of 0.0001, and a linear decay scheduler without warmup steps, using the AdamW [11] optimizer. The 64-dimensional model was trained for 280,000 steps (5,000 epochs), while the 128-dimensional model was trained for 56,000 steps (1,000 epochs). Following common practice in the literature, I used cross-entropy loss for training and evaluated top-1 and top-3 accuracy during training. The results are presented in Table \ref{tab: 1}.

\subsection{Downstream Task: Next Match Statistics Prediction (NMSP)}\label{subsec: Downstream Task: Next Match Statistics Prediction (NMSP)}
After pre-training, I replaced the MPP head with a multi-layer perceptron (MLP) designed to predict 18 team performance statistics (see Table \ref{tab: 3}) related to passing, offense, defense, and goalkeeping; these collectively provide a comprehensive description of team performance. Additionally, since the temporal positional encoding for NMSP was computed using 234 aggregated statistical variables (as detailed in Section 2.1), I removed the input MLP from the pre-training model and re-initialized its weights to accommodate the new input dimensions. The rest of the model's weights were initialized with the pre-trained MPP backbone. No input upsampling was applied, and the total 1,792 matches were used for fine-tuning. The data was split 80\% for training and 20% for validation.

During fine-tuning, all the model weights, including those loaded from the backbone, were updated. Each model (1 layer 64D and 1 layer 128D)  were trained with the same batch size of 256, using the same optimizer and learning rate scheduling strategy as in MPP, but with a warmup ratio of 0.1 and a weight decay of 0.01. These last two hyperparameters were used to prevent rapid changes in the pre-trained weights, which could lead to unstable fine-tuning. I used the average mean squared error (MSE) across all statistics as the training loss. 

Both models quickly converged after 2,000 training steps (~333 epochs). The training results are presented in Table \ref{tab: 2}, with more detailed evaluation scores presented in Table \ref{tab: 3}.

\subsection{Results}\label{subsec: Results}
The results presented in Table \ref{tab: 1} indicate that both transformer-based models with embedding dimensions of 64 and 128 perform effectively on the Masked Players Prediction (MPP) task, achieving comparable cross-entropy loss and top-1/top-3 accuracy. Notably, the 128-dimensional model converged faster, requiring fewer training steps to reach similar performance levels as the 64-dimensional model. The high top-3 accuracy across both models (>95\%) suggests that they are successfully capturing complex relationships between players, even with a relatively small dataset. These findings are promising, as they validate the chosen architecture and highlight the potential of using MPP to learn football player embeddings, compared to manually crafting features from raw statistics.

\begin{table}[t]
\caption{Performance of RisingBALLER on the Masked Players Prediction (MPP). The table presents the cross-entropy loss and top-1/top-3 accuracy scores for two transformer-based models with different embedding dimensions, 1l64d: 1 layer 64 embeddings dimension.}
\label{tab: 1}
\begin{center}
\begin{tabular}{cccccc}
\hline
\multicolumn{1}{c}{\bf Architecture}  &\multicolumn{1}{c}{\bf Split} &\multicolumn{1}{c}{\bf Steps} &\multicolumn{1}{c}{\bf Cross Entropy} & \multicolumn{1}{c}{\bf Accuracy} & \multicolumn{1}{c}{\bf Accuracy top3} 
\\\hline\\
1l64d &train & 280K & 0.3873 & x & x\\
1l64d &validation & 280k & \textbf{0.8434} & \textbf{0.7893} & \textbf{0.9537}\\
1l128d & train & 56K & 0.3236 & x & x\\
1l128d & validation & 56K & 0.8804 & 0.7764 & 0.9507 \\
\hline
\end{tabular}
\end{center}
\end{table}

Table \ref{tab: 2} showcases the overall performance of the RisingBALLER model on the Next Match Statistics Prediction (NMSP) task, benchmarking it against a standard baseline that predicts a team's next match performance by averaging the results of the previous five matches. The global metrics, including the mean squared error (MSE) and percentage improvement over the baseline, demonstrate that the transformer-based models with 64 and 128 embedding dimensions significantly outperform the baseline. Notably, the 128-dimensional model achieves a \textbf{37.70\%} improvement, while the 64-dimensional model shows a \textbf{35.35\%} improvement, underscoring the model's ability to generalize and capture complex patterns in football match data.

\begin{table}[t]
\caption{Global performance of the RisingBALLER models on the Next Match Statistics Prediction (NMSP), in comparison to a strong baseline method. The table reports the average mean squared error (MSE) across all statistics on both the training and validation splits, along with the percentage improvement over the baseline. The results demonstrate that the transformer-based model, particularly with 128-dimensional embeddings, significantly outperforms the baseline, confirming its effectiveness in modeling complex relationships in football data for statistics prediction.}
\label{tab: 2}
\begin{center}
\begin{tabular}{cccc}
\hline
\multicolumn{1}{c}{\bf Architecture}  &\multicolumn{1}{c}{\bf Split} &\multicolumn{1}{c}{\bf Global Average MSE} &\multicolumn{1}{c}{\bf \%improvement compared to baseline} 
\\\hline\\
baseline & validation & 819.28 & x \\
1l64d &train & 446.57 & x\\
1l64d &validation & 529.65 & 35.35\% \\
1l128d & train & 354.04 & x \\
1l128d & validation & 510.33 & \textbf{37.70\%} \\
\hline
\end{tabular}
\end{center}
\end{table}

Table \ref{tab: 3} provides a more granular analysis by presenting the root mean squared error (\textbf{rmse}) and dispersion coefficient \textbf{$\delta$} for each individual statistic. The dispersion coefficient is particularly important as it normalizes the rmse relative to the mean value of each statistic, making it a scale-independent measure of prediction accuracy. This allows for a more intuitive comparison across different statistics, regardless of their natural scale (e.g., the number of passes vs. the number of interceptions).

Also, I calculated the percentage difference based on the dispersion coefficient, showing how much the model improves or not compared to the baseline. For most statistics, RisingBALLER shows positive improvements, such as a 4\% better accuracy in predicting pass crosses and total shots with the 128-dimensional model. However, there are also a few areas, like expected goals (xG) and goals scored, where the model slightly underperforms compared to the baseline. This indicates that while the model is generally effective, there are certain aspects that could benefit from further refinement.

\begin{table}[t]
\caption{Detailed performance metrics for individual statistics in the NMSP task, comparing the baseline method with the RisingBALLER models (using 64-dimensional and 128-dimensional embeddings). The table presents the values for \textbf{rmse}$|$\textbf{$\delta$} for each statistic, as well as the percentage difference in the dispersion coefficient compared to the baseline. The results illustrate that RisingBALLER generally enhances prediction accuracy, especially in critical areas like total shots and pass crosses, while also identifying areas for further improvement in goal-related statistics.}
\label{tab: 3}
\begin{center}
\begin{tabular}{cccccc}
\hline
\multicolumn{1}{c}{\bf Statistics}  &\multicolumn{1}{c}{\bf baseline (b)} &\multicolumn{1}{c}{\bf 1l128d} &\multicolumn{1}{c}{\bf \%$\delta$ diff vs (b)} & \multicolumn{1}{c}{\bf 1l64d} & \multicolumn{1}{c}{\bf \%$\delta$ diff vs (b)} 
\\\hline\\
Pass total & $95.64|0.20$ & $\textbf{92.66}|\textbf{0.19}$ & \textcolor{blue}{+1} & $94.38|0.20$ & 0\\
Pass cross & 5.94$|$0.48 & \textbf{5.48$|$0.44} & \textcolor{blue}{+4} & 5.62$|$0.46 & \textcolor{blue}{+2}\\
Pass shot assist & 3.97$|$0.49 & \textbf{3.81$|$0.47} & \textcolor{blue}{+2} & 3.87$|$0.48 & \textcolor{blue}{+1}\\
Pass goal assist & \textbf{1.01$|$1.22} & 1.03$|$1.24 & \textcolor{red}{-2} & 1.07$|$1.29 & \textcolor{red}{-7}\\
Pass through ball & 2.37$|$1.08 & \textbf{2.33$|$1.08} & 0 & 2.32$|$1.08 & 0\\
Shot total & 5.23$|$0.42 & \textbf{4.95$|$0.38} & \textcolor{blue}{+4} & 4.97$|$0.40 & \textcolor{blue}{+2} \\ 
Shot xG & \textbf{0.79$|$0.64} & 0.90$|$0.73 & \textcolor{red}{-9} & 0.92$|$0.74 & \textcolor{red}{-10}\\
Shot goal & \textbf{1.27$|$1.05} & 1.34$|$1.09 & \textcolor{red}{-4} & 1.31$|$1.07 & \textcolor{red}{-2}\\
Interception won & \textbf{3.56$|$0.53} & 3.65$|$0.55 & \textcolor{red}{-2} & 3.78$|$0.57 & \textcolor{red}{-4}\\
Block won & 6.16$|$0.32 & 6.08$|$0.31 & \textcolor{blue}{+1} & \textbf{5.96$|$0.30} & \textcolor{blue}{+2}\\
Clearance total & 10.26$|$0.44 & 9.92$|$0.43 & \textcolor{blue}{+1} & \textbf{9.87$|$0.42} & \textcolor{blue}{+2}\\
Ball recovery total & 10.09$|$0.19 & \textbf{9.82$|$0.18} & \textcolor{blue}{+1} & 10.28$|$0.20 & \textcolor{red}{-1}\\
Counterpress total & \textbf{13.77$|$0.24} & 13.80$|$0.24 & 0 & 14.30$|$0.25 & \textcolor{red}{-1}\\
Dribble complete & \textbf{4.04$|$0.43} & 4.10$|$0.44 & \textcolor{red}{-1} & 4.08$|$0.44 & \textcolor{red}{-1}\\
Foul won total & 4.73$|$0.33 & \textbf{4.72$|$0.32} & \textcolor{blue}{+1} & 4.82$|$0.33 & 0\\
Foul committed total & \textbf{4.82$|$0.31} & 4.84$|$0.32 & \textcolor{red}{-1} & 4.96$|$0.33 & \textcolor{red}{-2}\\
Keeper save & 2.68$|$0.60 & \textbf{2.49$|$0.57} & \textcolor{blue}{+3} & 2.59$|$0.59 & \textcolor{blue}{+1}\\
Keeper shot saved & 4.19$|$0.51 & \textbf{4.09$|$0.50} & \textcolor{blue}{+1} & 4.14$|$0.50 & \textcolor{blue}{+1}\\
\hline
\end{tabular}
\end{center}
\end{table}

\section{RisingBALLER applications, the power of learnable embeddings}\label{sec: RisingBALLER applications, the power of learnable embeddings}
The embeddings learned by RisingBALLER can be leveraged to analyze player similarities, assess player-team compatibility, and provide tactical insights through spatial positional embeddings. In the following sections, I present an extensive analysis of these embeddings. Unless otherwise specified, the player and position embeddings used were obtained from the best-performing MPP model, 1l64d, as described in \ref{subsec: Results}.

\subsection{Spatial Positional Embeddings Analysis}\label{subsec: Spatial Positional Embeddings Analysis}
Accurately encoding player positions on the field is crucial in football analytics for tactical analysis and retrieving similar player profiles. Traditional methods often represent player positions using x and y coordinates or similar techniques. For instance, [7] encodes positions in a three-dimensional Euclidean space to ensure that distances between players in equivalent roles on opposite sides of the field (e.g., left back and right back) reflect their profile similarity. While these approaches provide a high degree of interpretability, they fail to capture the dynamic and evolving nature of modern football positioning.

In contemporary football, positions are defined not only by traditional spatial zones but also by the intrinsic attributes and dynamic roles players assume during matches. For example, in Chelsea's 2016-2017 Premier League title-winning campaign, Antonio Conte deployed Marcos Alonso, typically a left back, as a left wing-back, while Victor Moses, an attacking winger, operated as a right wing-back in a 3-4-3 formation. More recently, Pep Guardiola showcased positional fluidity by utilizing John Stones, a natural center-back, as a defensive and offensive midfielder in Manchester City's 2022 Champions League final. These examples illustrate how modern football positions are increasingly characterized by the versatility and adaptability of players, rather than rigid tactical zones.

To effectively capture this evolution, it is essential to reflect these nuanced roles in data-driven player performance analysis. RisingBALLER addresses this challenge by learning high-level positional embeddings that integrate both traditional spatial roles and player-specific attributes and behaviors observed during each match.

\subsubsection{Clustering Analysis of Positional Embeddings}\label{subsubsec: Clustering Analysis of Positional Embeddings}
\begin{figure}[h]
\begin{center}
\includegraphics[width=\textwidth]{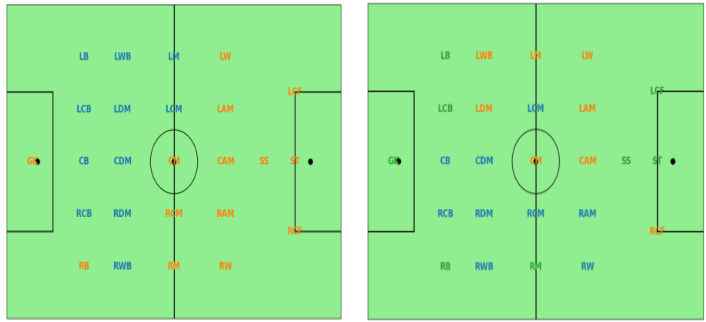}
\end{center}
\caption{Left: Positional embeddings clustered into 2 groups. Right: Positional embeddings clustered into 3 groups.}
\label{fig: 2}
\end{figure}

To assess whether the positional embeddings in RisingBALLER accurately represent spatial zones on the field, I conducted a clustering analysis using embeddings derived from match-level data. The analysis considered all 25 distinct positions defined in the StatsBomb data annotation format. By clustering these positions into two and three groups, I aimed to hierarchically examine the model's ability to differentiate between tactical zones. Figure \ref{fig: 2} illustrates the resulting clusters:
\begin{itemize}
    \item The two clusters (figure \ref{fig: 2}, left) clearly distinguish between defensive/midfield roles and attacking roles, which is expected in football's tactical organization. Notably, positions such as wing-backs (RWB, LWB) and defensive midfielders (LDM, CDM) are grouped with central defenders (CB, RCB, LCB), indicating that the model understands the importance of defensive contributions in these roles. Meanwhile, attacking players, including central attacking midfielders (CAM, RAM), wingers (RW, LW), and strikers (ST, SS), are clustered together, reflecting their forward-driven responsibilities.
    \item With three clusters (figure \ref{fig: 2}, right), the model captures additional nuances. For instance, the blue cluster predominantly contains central players (center-backs, central midfielders), who are responsible for maintaining defensive and midfield control, while the orange cluster includes more tactically flexible players, such as wide attackers and creative midfielders (LAM, LW, CAM). The green cluster, interestingly, groups players like full-backs (RB, LB), strikers (ST, SS), and goalkeepers (GK), reflecting their more isolated and specialized roles on the field.
\end{itemize}

\subsubsection{Positional Embeddings and Player Embeddings Similarity Analysis}\label{subsubsec: Positional Embeddings and Player Embeddings Similarity Analysis}
In addition to examining spatial zones, I evaluated whether RisingBALLER’s positional embeddings effectively capture players' intrinsic attributes and nuanced roles on the field. Using cosine similarity on the learned embeddings, I retrieved the top 10 most similar players for each position. This analysis focused on 1,945 players who participated in at least 10 matches in the dataset; the results revealed several interesting observations for some players.
\begin{itemize}
    \item \textbf{Yaya Touré (Manchester City)} was among the top 3 most similar players to the Central Attacking Midfielder (CAM) position, alongside Javier Pastore (Paris Saint-Germain) and Josip Iličić (ACF Fiorentina), despite being listed primarily as a defensive midfielder (CDM) in most starting lineups. This reflects how well RisingBALLER captures his dynamic role at City, where he frequently pushed forward, contributing creatively and offensively.
    \item \textbf{Neymar Da Silva (Barcelona)} emerged as the most similar player to the Right Central Forward (RCF) position, even though he predominantly played as a right winger (RW) in all matches. This suggests that the model captured his tendency to move inside and act as a secondary striker, aligning his playing influence more closely with a central forward role.
    \item \textbf{Antoine Griezmann (Atletico Madrid)}, traditionally positioned behind the main striker or as a wide forward or offensive midfielder, was identified as the most similar player to the Striker (ST) role, despite never being deployed as a conventional number nine. This highlights RisingBALLER's ability to recognize his impact in central attacking areas, especially during the 2015-2016 season.
\end{itemize}

These examples (more in Appendix \ref{appsubsec: Players and Positions Embeddings Similarity}) illustrate that RisingBALLER’s positional embeddings not only encode spatial zones but also capture the intrinsic qualities and adaptive roles of players during matches. The model’s understanding extends beyond conventional positional labels and rigid zonal classifications, accurately reflecting the tactical contributions of players like Yaya Touré, Neymar, and Griezmann. In the current era of data-driven performance analysis and recruitment, RisingBALLER offers a promising approach to representing modern football player positioning through data.

\subsection{Player Embeddings Analysis}
\label{subsec: Player Embeddings Analysis}
\subsubsection{Team Cohesion Estimation}
\label{subsubsec: Team Cohesion Estimation}
One of the primary priorities for football clubs regarding data analytics is player recruitment. In this context, clubs often generate shortlists of players that meet specific criteria by querying large databases. These queries typically rely on building similarity measures based on a wide range of statistics. Consequently, a key objective of the RisingBALLER architecture, particularly with the Masked Players Prediction (MPP) task, is to implicitly learn embeddings that cluster players with similar profiles. This process is analogous to how language models generate semantically meaningful embeddings in natural language processing. To assess the quality of the learned embeddings, I selected several players of interest and retrieved the 10 most similar players using cosine similarity (multiple examples are presented in Appendix \ref{appsubsec: Players Similarity Requests and Teams Cohesion estimation}, Table \ref{tab: 8}). This analysis yielded two key observations: \textbf{1. Players from the same team consistently appeared as the most similar players|Goalkeepers, regardless of their team affiliation, tended to cluster together more closely than outfield players.}

These findings provide intriguing insights. First, the model’s ability to learn similar embeddings for goalkeepers across different teams is likely attributed to the unique characteristics that distinguish them from outfield players. Unlike outfield players, whose roles can vary widely based on tactical systems and team context, goalkeepers share common fundamental tasks, which may allow the model to generalize effectively despite differences in team affiliations. The model captures the specific patterns associated with goalkeepers, resulting in embeddings that cluster them closely based on their shared functions on the field.

Conversely, the embeddings for outfield players are more heavily influenced by team affiliation. I think this phenomenon likely arises from the limited size of the dataset, which exposes the model to patterns relative to team identity rather than generalizing over player profiles across teams. Consequently, players from the same team tend to be embedded more closely together, especially when they occupy similar positions. While this indicates that the model successfully captures a team’s tactical identity by placing its players in a similar region, it also highlights a limitation in its ability to generalize player similarities across different teams for outfield roles.

Despite this limitation, the model's capacity to effectively represent teams by clustering their players together offers a way for quantifying team cohesion. Specifically, for all players in a squad, I calculate their cumulative similarity scores with all their teammates and average these scores by the squad size. The resulting value, which I call the “team cohesion factor” offers an overall measure of how well the players in a given team fit together. Table \ref{tab: 9} (Appendix \ref{appsubsec: Players Similarity Requests and Teams Cohesion estimation}) presents the 10 teams with the highest cohesion scores in the dataset alongside the 10 teams with the lowest scores. This metric could be useful for data-driven team building by accessing how much a player affects his team cohesion.

\subsubsection{Similar Player Retrieval}\label{subsubsec: Similar Player Retrieval}
Given that the model learns similar embeddings for players within the same team, it raises the hypothesis that including team affiliation embeddings (TE, section \ref{subsec: The Model Architecture}) as inputs for each player might be heavily contributing to this behavior. To test this hypothesis, I trained the two versions of the model 1l64d and 1l128d while ablating the team affiliation embeddings on MPP. Both models were trained for the same number of epochs (2,000) and on the same training and validation splits as previously, with the results reported in Table \ref{tab: 4}. In this setup, it was considerably harder for the model to predict the masked player, as evidenced by a significant drop in accuracy compared to the version with team affiliation embeddings (Table \ref{tab: 1}). This confirms the intuition that the team information provided as input makes the prediction task much easier, as the model tends to focus on patterns related to team identity.

Using the embeddings from the best-performing model without team affiliation embeddings (1l128d), I retrieved the 10 most similar players for a selection of players of interest and compared the quality of these retrievals to those obtained from the best-performing model with team affiliation embeddings (1l64d) (see Appendix \ref{appsubsec: Players Similarity Requests and Teams Cohesion estimation}, Table \ref{tab: 8}). The model that excluded team affiliation information demonstrated superior retrieval capabilities, as it focused on alternative patterns such as player positional attributes, raw statistics, and the frequency of adversarial matchups. This approach enabled it to learn similar embeddings for players with comparable profiles, independent of their team affiliations.

\begin{table}[t]
\caption{The MPP pre-training results without teams affiliation embeddings; The clear drop in the accuracies for the two models showcases how harder it’s for RisingBALLER to predict the masked players, by being forced to learn more difficult patterns for players embeddings with a small matches dataset.}
\label{tab: 4}
\begin{center}
\begin{tabular}{cccccc}
\hline
\multicolumn{1}{c}{\bf Architecture}  &\multicolumn{1}{c}{\bf Split} &\multicolumn{1}{c}{\bf Steps} &\multicolumn{1}{c}{\bf Cross Entropy} & \multicolumn{1}{c}{\bf Accuracy} & \multicolumn{1}{c}{\bf Accuracy top3} 
\\\hline\\
1l64d &train & 112K & 1.2663 & x & x\\
1l64d &validation & 112k & \textbf{1.8417} & \textbf{0.4550} & \textbf{0.8006}\\
1l128d & train & 112K & 0.5402 & x & x\\
1l128d & validation & 112K & 1.6708 & 0.5843 & 0.8830 \\
\hline
\end{tabular}
\end{center}
\end{table}

\subsubsection{Summary}\label{subsubsec: Summary}
The analysis of the players embeddings confirms the architecture design and the choice of Masked Player Prediction (MPP) as an effective task for learning high-level representations of football players. Unlike the intricate hand-crafted features that typically rely on extensive statistics aggregated per season for each player, often difficult to reproduce and scale, RisingBALLER provides a more streamlined alternative. It eliminates the need for multiple pre-processing steps, such as variable grouping, transformations, standardization, expert knowledge-based weighting, season-level aggregation, and rigid spatial position encoding that were employed before for football player representation [1,2]. Instead, RisingBALLER is built on a features learning method from raw match statistics and lineup data. The model benefits from a self-supervised training method, specifically masked player prediction, and utilizes a well-established representation learning architecture, the transformer, widely used in the literature [1,2,3,4,5,12], to directly encode players similarity at the match level, without requiring any expert-labeled data.

\section{Computational resources/Technical details}\label{sec: Computational resources/Technical details}
I used google colab pro for all the experiments in this research work. I did the RisingBALLER models training with the Hugging Face Trainer\footnote{https://huggingface.co/docs/transformers/main\_classes/trainer}, on 1 NVIDIA GPU Tesla T4. RisingBALLER’s codes for MPP and NMSP are written in PyTorch. The table \ref{tab: 5} presents the training times of the MPP pre-training experiments. For NMSP, each fine-tuning took between 5 and 10 minutes.

\begin{table}[t]
\caption{Masked Players Prediction training time.}
\label{tab: 5}
\begin{center}
\begin{tabular}{cc}
\hline
\multicolumn{1}{c}{\bf Training}  &\multicolumn{1}{c}{\bf Time} 
\\
\hline\\
MPP 1l64d , 280K steps (Tab \ref{tab: 1}) & 2 h 16 min 50 s\\
MPP 1l128d, 56k steps (Tab \ref{tab: 1}) & 35 min 10 s\\
MPP 1l64d , 112K steps (Tab \ref{tab: 4}) & 1 h 03 min 02 s\\
MPP 1l128d, 112k steps (Tab \ref{tab: 4}) & 1 h 15 min 02 s\\
\hline
\end{tabular}
\end{center}
\end{table}

\section{Conclusion and Future Works}\label{sec: Conclusion and Future Works}
In this paper, I introduced RisingBALLER, a transformer-based model designed to represent football players and matches by treating each match as a sequence of players. The model was trained using the Masked Player Prediction (MPP), which allowed it to learn high-level contextualized player representations. In addition, I proposed Next Match Statistics Prediction (NMSP) as a downstream task for RisingBALLER, demonstrating that the fine-tuned model outperforms a commonly used football forecasting baseline based on the average across previous matches.

The architecture of RisingBALLER trained with MPP have proven effective in learning rich representations for both players and positions. These representations have practical applications in modern data-driven football analytics. They include accurate positional representation for tactical analysis, team cohesion estimation and similar player retrieval for better and faster recruitment.

\textbf{Future Work:} The architecture of RisingBALLER allows further applications, taking advantage of the model's flexibility and predictive power. They include for example:
\begin{itemize}
    \item \textbf{Conditioned Optimal Squad Generation (SQUADGEN):} one of the most promising future directions is to leverage the NMSP task to create optimal squads based on specific criteria (e.g., maximizing the number of crosses and shots). By predicting how different player lineups might perform in terms of offensive or defensive metrics, RisingBALLER could assist coaches and analysts in selecting squads that align with their tactical objectives.
    \item \textbf{Player Interactions Estimation:} The attention matrices within the transformer architecture provide a new method to model player interactions during matches. By analyzing the matrices from one or multiple attention heads, it is possible to extract insights regarding offensive, defensive, and passing synergies between players. This analysis could also reveal emerging interactions within the game.
    \item \textbf{Scaling-from RisingBALLER to a fully accomplished BALLER:} Language models have improved primarily through scaling the transformer architecture used in BERT [2] to larger models [12]. This scaling has led to the emergence of new capabilities in larger models pretrained on extensive text datasets, resulting in the well-known large language models (LLMs) we see today. By increasing the size of the RisingBALLER model, expanding the player vocabulary, and leveraging larger datasets with a greater diversity of matches, could we anticipate the development of a comprehensive football model that demonstrates a high-level understanding of all aspects of the game, including player positions, team identity, and unique playing styles?
\end{itemize}

\subsubsection*{Acknowledgments}
This paper will be presennted at the StatsBomb Conference 2024\footnote{https://statsbomb.com/events/statsbomb-conference-2024/}, Research Stage. I would like to thank StatsBomb for providing the open data used to train the models, and I am especially grateful to my advisors from StatsBomb, Francisco Goitia and Duncan Hunter, for their valuable feedbacks.

\newpage
\section*{REFERENCES}
[1] Vaswani, A., Shazeer, N., Parmar, N., Uszkoreit, J., Jones, L., Gomez, A., Kaiser, Ł., Polosukhin, I. (2017). Attention is All You Need. In Proceedings of the 31st Conference on Neural Information Processing Systems (NeurIPS 2017). https://arxiv.org/abs/1706.03762

[2] Devlin, J., Chang, M. W., Lee, K., \& Toutanova, K. (2018). BERT: Pre-training of Deep Bidirectional Transformers for Language Understanding. In Proceedings of the 2018 Conference of the North American Chapter of the Association for Computational Linguistics (NAACL 2018). https://arxiv.org/abs/1810.04805

[3] Liu, Y., Ott, M., Goyal, N., Du, J., Cardie, C., \& Gardner, M. (2019). RoBERTa: A Robustly Optimized BERT Pretraining Approach. In Proceedings of the 2019 Conference of the North American Chapter of the Association for Computational Linguistics (NAACL 2019). https://arxiv.org/abs/1907.11692

[4] Radford, A., Kim, J. W., Xie, L., et al. (2021). Learning Transferable Visual Models From Natural Language Supervision. In Proceedings of the 2021 IEEE/CVF International Conference on Computer Vision (ICCV 2021). https://arxiv.org/abs/2103.00020

[5] Caron, M., Touvron, H., Misra, I., et al. (2021). Emerging Properties in Self-Supervised Vision Transformers. In Proceedings of the 2021 IEEE/CVF International Conference on Computer Vision (ICCV 2021). https://arxiv.org/abs/2104.14294

[6] FIFA. The FIFA Football Language

[7] Akhanli, S., Distance construction and clustering of football player performance data (2019), Thesis for PhD.

[8] Dinsdale, D., Gallagher, J., (2022). Transfer Portal: An Efficient Framework for Transfer Learning. In Proceedings of the 2022 Conference on Neural Information Processing Systems (NeurIPS 2022). https://arxiv.org/abs/2201.11533

[9] McHale, I., Holmes, B., Estimating Fee Transfers of Football Players using advanced performance metrics and machine learning. European Journal of Operational Research.

[10] Trower, M., Graham, N., Cottrell, N., Hengster, Y., (2023). Clustering Women’s Football Players: Identifying Functional Patterns for Performance Optimization. StatsBomb Conference 2023 research papers.

[11] Loshchilov, I., \& Hutter, F. (2017). AdamW: Decoupled Weight Decay Regularization. In Proceedings of the 2017 International Conference on Learning Representations (ICLR 2017). https://arxiv.org/abs/1711.05101

[12] Brown, Tom B., et al. (2020). Language Models are Few-Shot Learners. Advances in Neural Information Processing Systems (NeurIPS 2020). https://arxiv.org/abs/2005.14165

\newpage
\appendix
\section{Further experiments}\label{appsec: Further experiments}
\subsection{MPP Architecture search}\label{appsubsec: MPP Architecture search}
I conducted a comprehensive architecture search for Masked Player Prediction (MPP) within the RisingBALLER framework to identify the optimal model configuration for learning player representations. This study explored various transformer architectures by varying the number of layers and embedding dimensions. Each model was trained for 56,000 steps (1,000 epochs) under consistent training conditions with the same parameters as in section \ref{subsec: Pre-Training with Masked Players Prediction (MPP)}. The results presented in Table \ref{tab: 6} showed that models with two layers did not offer significant advantages over single-layer models, likely due to the small size of the used dataset. Notably, the single-layer model with 128-dimensional embeddings achieved the best performance in terms of validation loss and accuracy. Additionally, the 64-dimensional single-layer model also showed promising results, which led me to extend its training for an additional 4,000 epochs as presented before in Table \ref{tab: 1}.

\begin{table}[t]
\caption{Performance Metrics for Various Transformer Architectures in Masked Player Prediction (MPP). The table presents results for different model configurations, varying by the number of transformer layers (l) and embedding dimensions (d). For each architecture, we list the number of parameters of the model, training steps, cross-entropy loss on the validation set, and both top-1 and top-3 accuracy.}
\label{tab: 6}
\begin{center}
\begin{tabular}{cccccc}
\hline
\multicolumn{1}{c}{\bf Architecture}  &\multicolumn{1}{c}{\bf Split} &\multicolumn{1}{c}{\bf Steps} &\multicolumn{1}{c}{\bf Cross Entropy} & \multicolumn{1}{c}{\bf Accuracy} & \multicolumn{1}{c}{\bf Accuracy top3} 
\\\hline\\
1l32d & 185K & 56K & 1.2744 & 0.5537 & 0.8535\\
\textbf{1l64d} & 387K & 56K & \textbf{0.9262} & \textbf{0.7207} & \textbf{0.9355}\\
\textbf{1l128d} & 851K & 56K & \textbf{0.8804} & \textbf{0.7764} & \textbf{0.9507}\\
2l32d & 195K & 56K & 1.2819 & 0.5500 & 0.8524\\
2l64d & 427K & 56K & 0.9872 & 0.6728 & 0.9177\\
\hline
\end{tabular}
\end{center}
\end{table}

\subsection{MPP Pre-Training Ablation}\label{appsubsec: MPP Pre-Training Ablation}
To evaluate the impact of MPP pretraining, I fine-tuned the two transformer architectures 1l64d and 1l128d pre-trained on MPP for 56K steps for the architecture search on NMSP and also trained to the same architecture  from scratch on NMSP, under the same training conditions. The results are presented in Table \ref{tab: 7}.

\begin{figure}[h]
\begin{center}
\includegraphics[width=\textwidth]{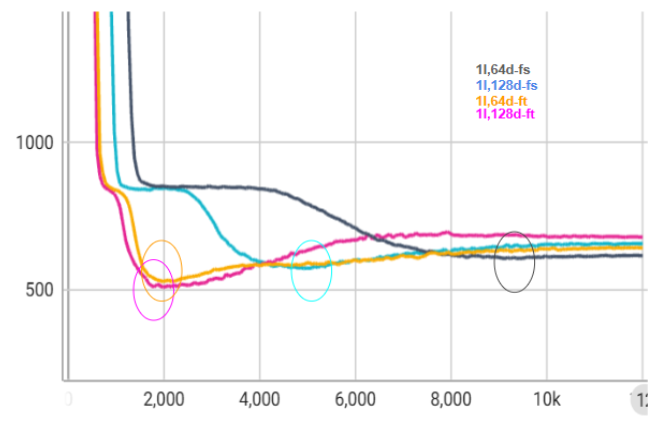}
\end{center}
\caption{Validation curves of the two architectures on NMSP, with and without MPP pretraining, fs means from scratch and ft means fine tuned. The oval circles denote the areas of convergence with the minimal losses.}
\label{fig: 3}
\end{figure}

The evaluation losses in figure \ref{fig: 3} show that the pretraining significantly enhances performance, a well-known transformer property. Specifically, the 1l,64d model fine-tuned with MPP achieved an evaluation loss of 529.65 representing 13\% of improvements compared to his from scratch version, and the fine-tuned 1l128d model archives 11\% of improvements compared to his from scratch version. Furthermore the validation curves in figure 3 show that models trained from scratch not only take longer to converge but also fall short of the performance levels achieved by the pre-trained models.

This ablation confirms that MPP pretraining is crucial for improving model effectiveness. Also, as the dataset and models size scale up, the benefits of the pretraining should likely be even more pronounced.

\begin{table}[t]
\caption{Evaluation Loss Comparison for NMSP with and without MPP Pre-training: the values are the global average MSE over all statistics.}
\label{tab: 7}
\begin{center}
\begin{tabular}{cccc}
\hline
\multicolumn{1}{c}{\bf Architecture} &\multicolumn{1}{c}{\bf Steps} &\multicolumn{1}{c}{\bf Eval loss with 
MPP pretraining} & \multicolumn{1}{c}{\bf Eval loss while from scratch}
\\\hline\\
1l, 64d & 414K & \textbf{529.65} & 607.29\\
1l, 128d & 909K & \textbf{510.33} & 574.33\\
\hline
\end{tabular}
\end{center}
\end{table}

\section{Embeddings Analysis}\label{appsec: Embeddings Analysis}
\subsection{Players and Positions Embeddings Similarity}\label{appsubsec: Players and Positions Embeddings Similarity}
\begin{figure}[h]
\begin{center}
\includegraphics[width=\textwidth]{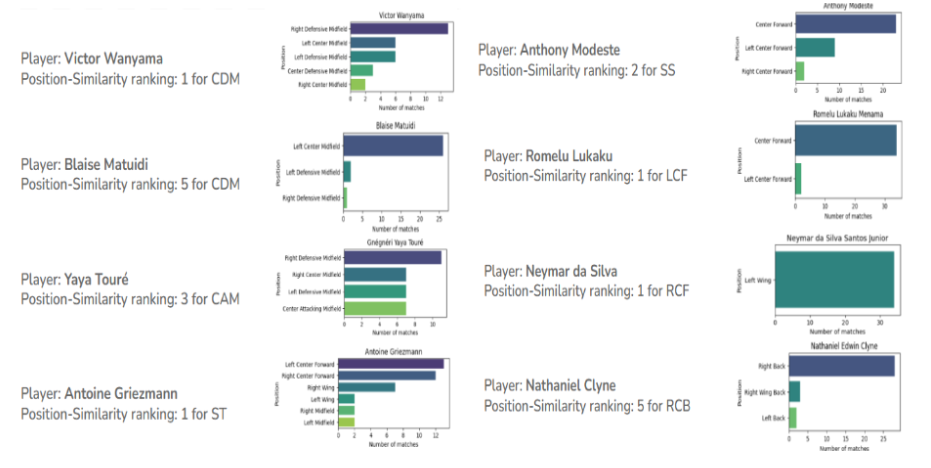}
\end{center}
\caption{}
\label{fig: 4}
\end{figure}

Figure \ref{fig: 4}: Example of nuanced positional embeddings learned by the model, highlighting the effective roles of players on the field. For each player, I show their ranking among 1,945 players (with at least 10 matches) based on similarity requests between their embeddings and the position embeddings. The bar plot shows the frequency of their native positions as reported in the StatsBomb dataset. All reported players are actually suited for positions different from their native ones, showcasing RisingBALLER's ability to learn high-level player positioning representations. C: central, D: defensive, M: midfield, A: attacking, ST: striker, SS: second striker, L: left, R: right, B: back, F: forward.

\subsection{Players Similarity Requests and Teams Cohesion estimation}\label{appsubsec: Players Similarity Requests and Teams Cohesion estimation}
Table \ref{tab: 8}: Comparison of the top 10 most similar players for selected key players from the 2015-2016 season, as retrieved by two versions of the model: one with team affiliation embeddings and one without. The model with team affiliation embeddings tends to group players from the same team, while the model without team information highlights players with similar attributes and playing styles across different teams. For example, when team affiliation is excluded, Cristiano Ronaldo is identified as similar to Neymar and Luis Suárez, indicating that the model captures positional and performance similarities more effectively without team bias. For goalkeepers, the model with teams affiliation seems to learn more diverse embeddings while the model without is too much biased to the league; a proof that having teams affiliation adds more complexity and discriminative features in goalkeepers representations. These examples illustrate the model’s capability to learn good features for individual player profiles through MPP, which can be valuable for data-driven player recruitment and squad optimization-(RM): Real Madrid, (B): Barcelona, (Ars): Arsenal, (ToT): Tottenham Hotspur, (Bay M): Bayern Munich, w: with, w/o: without, TE: team affiliation embeddings.

Table \ref{tab: 9}: Lists of the top 10 teams with the highest and lowest cohesion scores based on the RisingBALLER model’s player embeddings. The “team cohesion factor” is calculated by averaging the cumulative similarity scores of each player with all their teammates. Teams with high cohesion scores, such as Hannover 96 and Leicester City, indicate a strong alignment in player profiles, suggesting a well-integrated squad. In contrast, teams with low cohesion scores, such as Manchester United and AS Roma, have less alignment in player profiles, potentially reflecting tactical mismatches or a need for improved squad balance. \textbf{However}, assessing the effectiveness of this metric will need some football club expert knowledge. 

\begin{table}[h]
\centering
\caption{}
\label{tab: 8}
\begin{tabularx}{\textwidth}{X|X|X}
\toprule
\textbf{Player of interest} & \textbf{Top 10 most similar players w TE} & \textbf{Top 10 most similar players w/o TE} \\ 
\midrule
Cristiano Ronaldo (RM) & 
\begin{tabular}[c]{@{}l@{}}
Jesé Rodríguez Ruiz (RM) \\
Francisco Román Isco (RM) \\
Mateo Kovačić (RM) \\
James Rodríguez (RM) \\
Karim Benzema (RM) \\
Borja Mayoral Moya (RM) \\
Marcos Llorente Moreno (RM) \\
Gareth Frank Bale (RM) \\
Denis Cheryshev (Valencia) \\
Nacho Fernández (RM)
\end{tabular} & 
\begin{tabular}[c]{@{}l@{}}
Karim Benzema (Real Madrid) \\
Neymar da Silva (Barcelona) \\
Jesé Rodríguez (Real Madrid) \\
Danilo da Silva (Real Madrid) \\
Toni Kroos (Real Madrid) \\
Imanol Arruti (Real Sociedad) \\
Luis Suárez (Barcelona) \\
Dani Ceballos (Real Betis) \\
Gareth Bale (Real Madrid) \\
Nacho Fernández (Real Madrid)
\end{tabular} \\ 
\midrule
Lionel Messi (B) & 
\begin{tabular}[c]{@{}l@{}}
Arda Turan (B) \\
Munir El Haddadi Mohamed (B) \\
Sandro Ramírez Castillo (B) \\
Luis Alberto Suárez Díaz (B) \\
Sergi Roberto Carnicer (B) \\
Aleix Vidal Parreu (B) \\
Rafael Alcântara (B) \\
Neymar da Silva Santos(B) \\
Sergio Busquets i Burgos (B) \\
Sergi Samper Montaña (B)
\end{tabular} & 
\begin{tabular}[c]{@{}l@{}}
Munir El Haddadi (Barcelona) \\
Luis Suárez (Barcelona) \\
Deyverson Acosta (Levante UD) \\
Sandro Ramírez (Barcelona) \\
Santi Mina (Valencia) \\
Marco Asensio (Espanyol) \\
Marc Bartra (Barcelona) \\
Arda Turan (Barcelona) \\
Xabier Prieto (Real Sociedad) \\
Gareth Frank Bale (Real Madrid)
\end{tabular} \\ 
\midrule
Mesut Özil (Ars) & 
\begin{tabular}[c]{@{}l@{}}
Joel Nathaniel Campbell (Ars) \\
Francis Joseph Coquelin (Ars) \\
Mathieu Flamini (Ars) \\
Aaron Ramsey (Ars) \\
Jack Wilshere (Ars) \\
Santiago Cazorla (Ars) \\
Alex Iwobi (Ars) \\
Petr Čech (Ars) \\
Theo Walcott (Ars) \\
Per Mertesacker (Ars)
\end{tabular} & 
\begin{tabular}[c]{@{}l@{}}
Mathieu Flamini (Arsenal) \\
Aaron Ramsey (Arsenal) \\
Ross Barkley (Everton) \\
Yaya Touré (Manchester City) \\
Mohamed Elneny (Arsenal) \\
Gylfi Sigurðsson (Swansea City) \\
Sandro Cordeiro (West B. Albion) \\
Jack Wilshere (Arsenal) \\
Steven Davis (Southampton) \\
Darren Fletcher (West B. Albion)
\end{tabular} \\ 
\midrule
Kyle Walker (ToT) & 
\begin{tabular}[c]{@{}l@{}}
Kieran Trippier (ToT) \\
Danny Rose (ToT) \\
Clinton Mua N'Jie (ToT) \\
Harry Kane (ToT) \\
Nacer Chadli (ToT) \\
Erik Lamela (ToT) \\
Heung-Min Son (ToT) \\
Josh Onomah (ToT) \\
Ryan Mason (ToT) \\
Alassane Touré (GFC Ajaccio)
\end{tabular} & 
\begin{tabular}[c]{@{}l@{}}
Héctor Bellerín (Arsenal) \\
Joel Ward (Crystal Palace) \\
Daryl Janmaat (Newcastle Utd) \\
Kieran Trippier (T. Hotspur) \\
Cédric Soares (Southampton) \\
Branislav Ivanović (Chelsea) \\
Craig Dawson (West B. Albion) \\
Àngel Zaragoza (Swansea City) \\
James Tomkins (West Ham Utd) \\
Glen Johnson (Stoke City)
\end{tabular} \\ 
\midrule
Manuel Neuer (Bay M) & 
\begin{tabular}[c]{@{}l@{}}
Sven Ulreich (Bay M) \\
Jean-Louis Leca (Bastia) \\
Alban Lafont (Toulouse) \\
Stefano Sorrentino (Palermo) \\
Morgan De Sanctis (AS Roma) \\
Jack Butland (Stoke City) \\
Loris Karius (FSV Mainz 05) \\
Antonio Mirante (Bologna) \\
Gigio Donnarumma (AC Milan) \\
Stéphane Ruffier (Saint-Étienne)
\end{tabular} & 
\begin{tabular}[c]{@{}l@{}}
Oliver Baumann (Hoffenheim) \\
Roman Bürki (B. Dortmund) \\
Felix Wiedwald (Werder Bremen) \\
Christian Mathenia (Darmstadt 98) \\
Lukáš Hrádecký (E. Frankfurt) \\
Ralf Fährmann (Schalke 04) \\
René Adler (Hamburger SV) \\
Yann Sommer (B. Gladbach) \\
Przemysław Tytoń (VfB Stuttgart) \\
Loris Karius (FSV Mainz 05)
\end{tabular} \\ 
\bottomrule
\end{tabularx}
\end{table}

\begin{table}[h]
\centering
\caption{}
\label{tab: 9}
\begin{tabularx}{\textwidth}{X|X}
\toprule
\textbf{Top 10 Teams with the Highest Cohesion Scores} & \textbf{Top 10 Teams with the Lowest Cohesion Scores} \\ 
\midrule
Hannover 96 (Bundesliga) & AS Monaco (Ligue 1) \\ 
Ingolstadt (Bundesliga) & VfB Stuttgart (Bundesliga) \\ 
Darmstadt 98 (Bundesliga) & Everton (Premier League) \\ 
Leicester City (Premier League) & Sunderland (Premier League) \\ 
FC Köln (Bundesliga) & Newcastle United (Premier League) \\ 
Gazélec Ajaccio (Ligue 1) & Sevilla (La Liga) \\ 
Real Madrid (La Liga) & Manchester United (Premier League) \\ 
RC Deportivo La Coruña (La Liga) & Lille (Ligue 1) \\ 
Napoli (Serie A) & AS Roma (Serie A) \\ 
Caen (Ligue 1) & Troyes (Ligue 1) \\ 
\bottomrule
\end{tabularx}
\end{table}

\subsection{Analysis of Players Dissimilarity heatmap, the case of Barcelona vs Real Madrid}\label{appsubsec: Analysis of Players Dissimilarity heatmap, the case of Barcelona vs Real Madrid}

\begin{figure}[h]
\begin{center}
\includegraphics[width=\textwidth]{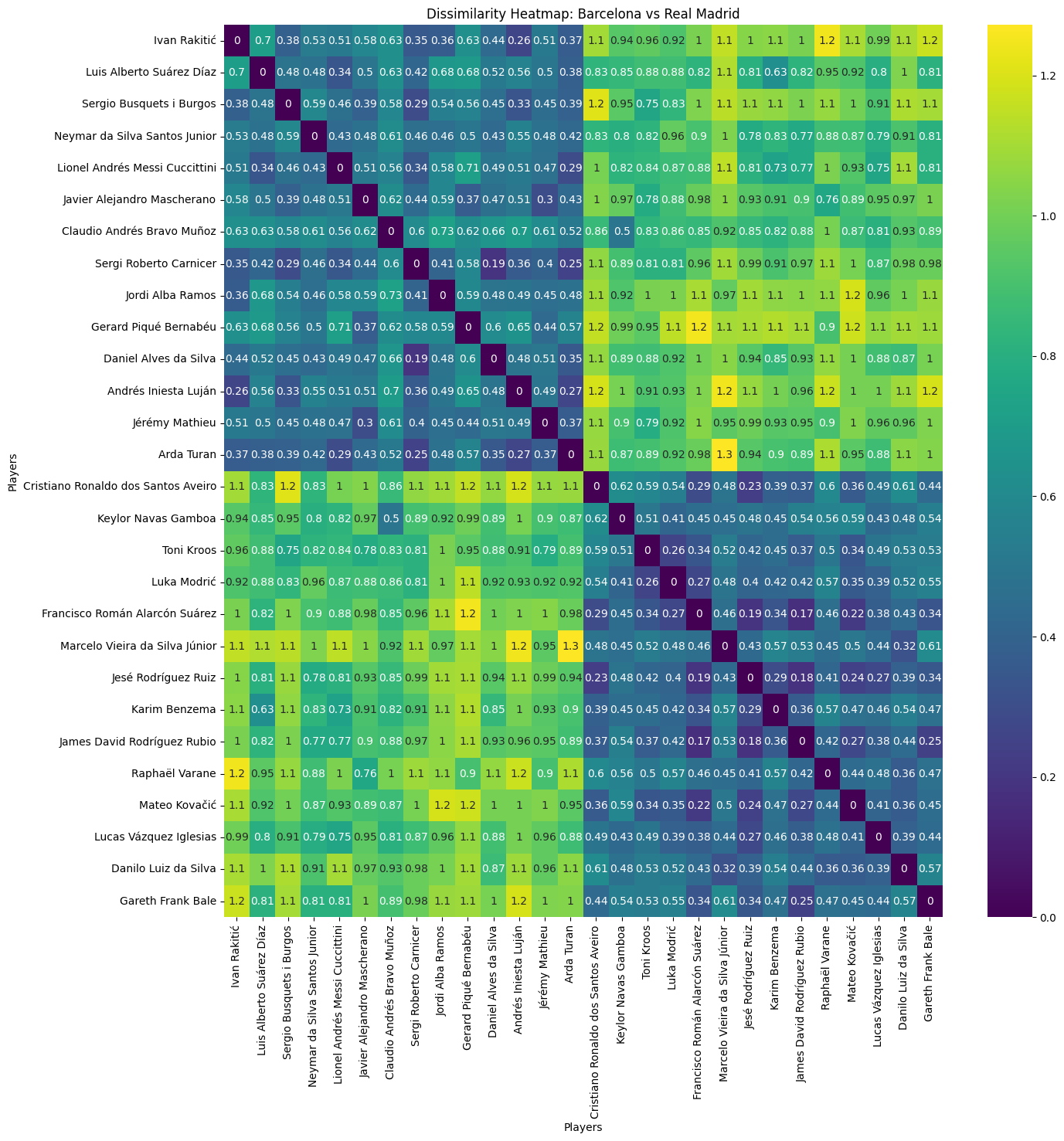}
\end{center}
\caption{Dissimilarity heatmap with players affiliation embeddings, the players embeddings are compared using cosine similarity, the players embeddings used are from the best performing model on MPP in that setup, 1l64d (scores in Table \ref{tab: 1}).}
\label{fig: 5}
\end{figure}

I created two heatmaps showing the dissimilarity (lower is better) between the 14 most frequent players from FC Barcelona and Real Madrid in the dataset. The first heatmap (figure \ref{fig: 5}) shows that when team affiliation embeddings are used in the model, players from the same team have very low dissimilarity, forming clear blocks within each team's matrix, while the cross-team matrix has high dissimilarity values. This indicates that the model is heavily relying on team identity to form its player embeddings.

\begin{figure}[h]
\begin{center}
\includegraphics[width=\textwidth]{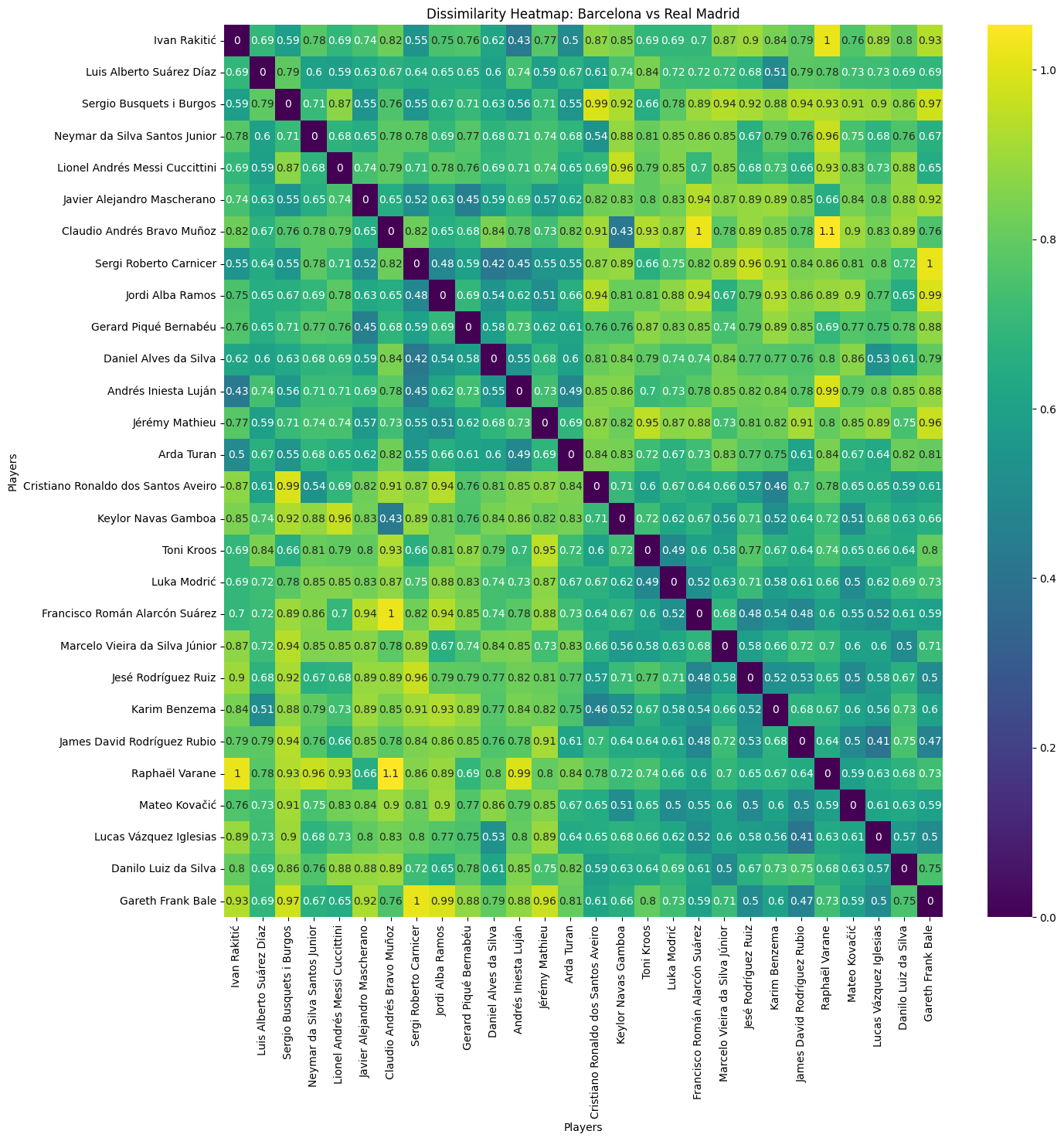}
\end{center}
\caption{Dissimilarity heatmap without players affiliation embeddings, the players embeddings are compared using cosine similarity, the players embeddings used are from the best performing model on MPP in that setup, 1l128d (scores in Table \ref{tab: 4}).}
\label{fig: 6}
\end{figure}

In contrast, the second heatmap (figure \ref{fig: 6}), generated from the model without team affiliation embeddings, shows that the model focuses more on individual player characteristics, learning similar embeddings for players with similar profiles, even if they play for opposing teams. For instance, Karim Benzema and Luis Suárez, both top-tier central forwards, show greater similarity. Likewise, Neymar shows more similarity to Cristiano Ronaldo, and Lionel Messi is more similar to Gareth Bale. Midfielders like Andrés Iniesta, Toni Kroos, and Luka Modrić also form more similar groups, and defenders like Raphael Varane and Gerard Piqué share closer embeddings.

\end{document}